\documentclass[letterpaper, 10 pt, conference]{ieeeconf}  

\IEEEoverridecommandlockouts                              

\usepackage{mathtools}
\usepackage{amsmath}
\usepackage{graphics}
\usepackage{caption}
\usepackage{subcaption}
\usepackage{amssymb}
\usepackage{ctable}
\usepackage{cite}
\usepackage{setspace}
\usepackage{bm}
\usepackage{svg}

\usepackage[noend]{algpseudocode}
\usepackage{algorithmicx,algorithm}

\usepackage{xcolor}
\usepackage{tabularx}
\usepackage[short]{optidef}
\usepackage{url}

\usepackage{colortbl,booktabs}
\usepackage{multirow}
\usepackage{multicol}

\usepackage{hyperref}
\hypersetup{
  bookmarksnumbered=true,
  bookmarksopen=true,
            colorlinks=false,
            bookmarksnumbered=true,
            linkcolor=blue,
            anchorcolor=blue,
            citecolor=green}

\title{\LARGE \bf
Simulation Platform for Autonomous Aerial Manipulation in Dynamic Environments
}

\author{Fengyu Quan$^{1}$, Huisheng Huang$^{1}$, Hongjie Zeng$^{1}$, Haoyao Chen$^{1}$, and Yunhui Liu$^{2}$
\thanks{* This work was partially supported by grants from the National Natural Science Foundation of China (Reference Nos. 61673131 and U1713206).}
\thanks{$^{1}$ Department of Mechanical Engineering and Automation in Harbin Institute of Technology Shenzhen, and the State Key Laboratory of Robotics and System, China.
Corresponding author:
        {\tt\small hychen5@hit.edu.cn}}%
\thanks{$^{2}$ Department of Mechanical and Automation Engineering in Chinese University of Hong Kong, Hong Kong, P. R. China. {\tt\small yhliu@mae.cuhk.edu.hk}}
}

\begin{document}

\maketitle
\thispagestyle{empty}
\pagestyle{empty}

\maketitle

\begin{abstract}

The aerial manipulator (AM) is a systematic operational robotic platform in high standard on algorithm robustness. Directly deploying the algorithms to the practical system will take numerous trial and error costs and even cause destructive results. In this paper, a new modular simulation platform is designed to evaluate aerial manipulation related algorithms before deploying. In addition, to realize a fully autonomous aerial grasping, a series of algorithm modules consisting a complete workflow are designed and integrated in the simulation platform, including perception, planning and control modules. This framework empowers the AM to autonomously grasp remote targets without colliding with surrounding obstacles relying only on on-board sensors. Benefiting from its modular design, this software architecture can be easily extended with additional algorithms. Finally, several simulations are performed to verify the effectiveness of the proposed system.

\end{abstract}


%
\IEEEpeerreviewmaketitle

\section{Introduction}

In the past decade, many researchers have focused on the active intersection between micro aerial vehicle (MAV) and the external environments. 
However, since the MAVs may easily crash due to immature algorithms, conducting a real world experiment is time-consuming and expensive. To make it more convenient to test and enhance the success rate of experiments, developing a complete aerial simulation framework is of urgent need.

A flight simulator is required to simulate real physical characteristics as much as possible, including gravity, light, collision and so on. Gazebo \cite{noori20173d} is widely used in multi-robot simulation; physical models, sensors, and actuators, etc. are allowed to add and establish a simulated physical environment similar with a real world. 
There already exist multiple gazebo-based MAV simulation platforms like Hector Quadrotor\cite{hector_quarotor} and RotorS \cite{furrer2016rotors}. They are both designed for physical simulations of MAVs, and are able to provide the evaluation of complicated algorithms like simultaneous localization and mapping(SLAM) and path planning. 
Based on RotorS, Antonio et al. \cite{loquercio2019deep} realized the drone racing simulation and migrated it to real world experiments.
AirSim \cite{shah2018airsim} constructed a hi-fidelity physical world based on the Unreal Engine \cite{sanders2016introduction}, aiming to construct a training set under different conditions for the research of reinforcement learning; it is claimed that the real world experiments match well with the simulations, confirming the effectiveness and necessity of the simulator.
Other simulators like Morse \cite{echeverria2011modular} and jMavSim \cite{pixhawk} are also widely used. Except the physical engine-based simulators, other works for MAVs are generally developed based on numerical platform Matlab/Simulink \cite{suarez2017lightweight}. 
However, all the above MAV platforms are developed without considering the evaluation ability of autonomous aerial manipulation in which diverse behaviors need to be contained.

\begin{figure}[t]
    \centering
    \includegraphics[width = \linewidth]{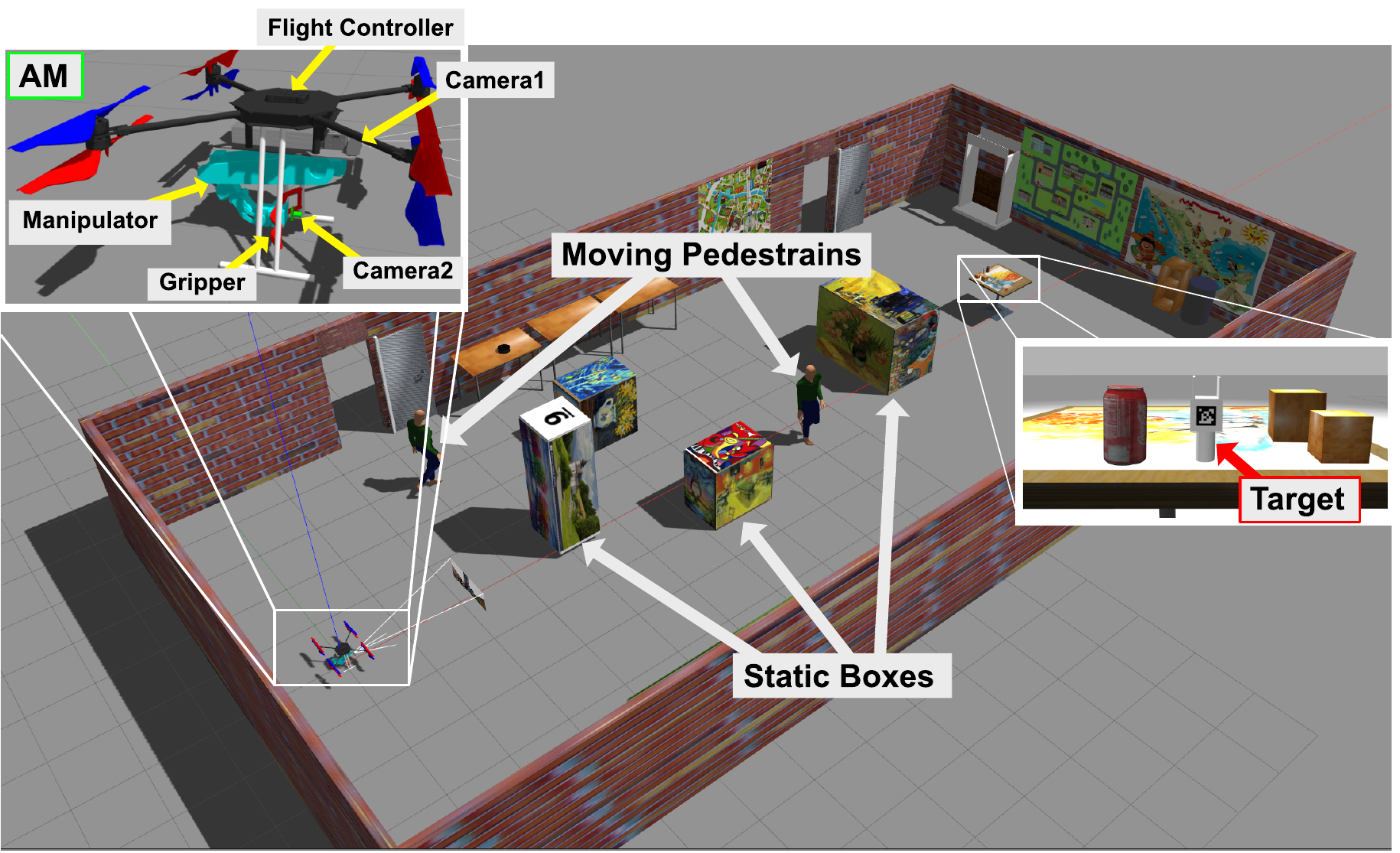}
    \caption{Simulated dynamic environment for remote autonomous aerial manipulation.}
    \label{fig:simulation_environmnet}
\end{figure}

\par In terms of aerial manipulation, aerial grasping is one of the most important capabilities.
Considering the system power-weight ratio, attaching a single high DOF manipulator on the aerial platform is an economical solution, and is able to achieve a large working space to complete complex manipulation tasks.
To fulfill stable grasping operation during the flight, existing works mainly focus on eliminating the coupling force between the MAV and manipulator \cite{kim2017robust}, reducing the disturbance caused by contacting with the operating environment \cite{meng2019hybrid}, and using visual feedback information to achieve grasping of high success rate \cite{lippiello2018image}. 
However, these results generally depend on external localization systems to supply pose feedback and the clear surrounding environments, while such conditions can not always be guaranteed in many practical applications. 
On the other hand, according to our knowledge, in current related work, the target to be grasped is prone to be perceived easily at the beginning instead of being occluded by obstacles or out of sight.
Conditions mentioned above are all considered in our developed simulation platform and a set of algorithms are integrated to tackle these difficult problems.

\par The contributions of this paper are two-fold.

First, a modular simulation platform is developed by combining software and hardware models. Modules of online perception, path planning, visual servoing controller, and Pixhawk\cite{meier2011pixhawk}-based flight controller for aerial manipulation are developed and integrated into the platform, allowing users to test a complete autonomous grasping process.

Second, a novel aerial manipulating framework is proposed to realize an autonomous remote grasping in cluttered dynamic scenarios. The proposed approach only relies on onboard sensors without external localization systems, and considers dynamic obstacles existing on the pre-planned path. Our developed framework has been successfully verified in our simulation platform, and the whole system will be open source in the near future.

\section{Simulation Platform Design}

\par Autonomous grasping in dynamic environments is a complicated job, and thus need to contain diverse modules including perception, planning and control, etc. Examining the feasibility and robustness of algorithms is challenging in real world environments.  
The design of a modular simulator is therefore of vital importance, and the simulator should allow users to design and verify their algorithms before applied to the real robotic platforms.
To reduce the gap from simulations to real scenes, three basic requirements should be satisfied, described as follows:

1) Both static and dynamic obstacles should be contained in the environment;

2) The grasp target is far away from the take-off spot, and only the approximate region other than the exact location is known in advance due to occlusion or sight limitation;

3) The localization of the aerial manipulator itself only relies on on-board sensors. 

By considering above requirements, a new physical simulation platform for aerial manipulation is proposed based on ROS and Gazebo. As shown in Fig. \ref{fig:simulation_environmnet}, the platform builds a physical world with static boxes and dynamic pedestrians, and contains an octocopter with a 6-DOF lightweight manipulator. An RGBD camera fixed on the aerial base link is used to perceive surrounding environment, while another camera is mounted at the end of the manipulator to track the grasp target. 

To realize aerial exploration, grasping and transportation, the proposed software framework consists of three fundamental modules:

1) The perception module utilizes the camera mounted on the aerial base to perceive surrounding obstacles and the camera on the end-effector to identify the target of interest. At the same time, the state estimation of the aerial manipulator itself, with respect to the inertial coordinate frame, is determined based on the visual information. The state estimation here is also defined as the well-known visual SLAM problem.

2) The planning module consists of the global and local planners as well as the target searching planner. The global planner firstly generates a feasible path between the starting location and the end location; the planning is performed in a pre-built voxel map to prevent from falling into the local minimum. The local planner makes necessary modifications along the global path to avoid dynamic obstacles during approaching and grasping procedures. And the path generation for searching target objects is also needed, if the target is not found when the system arrives at the end location.

3) The control module consists of the high-level and low-level controllers. The high-level controller consists of a visual servoing controller and path following controller, while the low-level controller drives the aerial manipulator to the desired state. 

\begin{figure}[t]
    \centering
    \includegraphics[width = \linewidth]{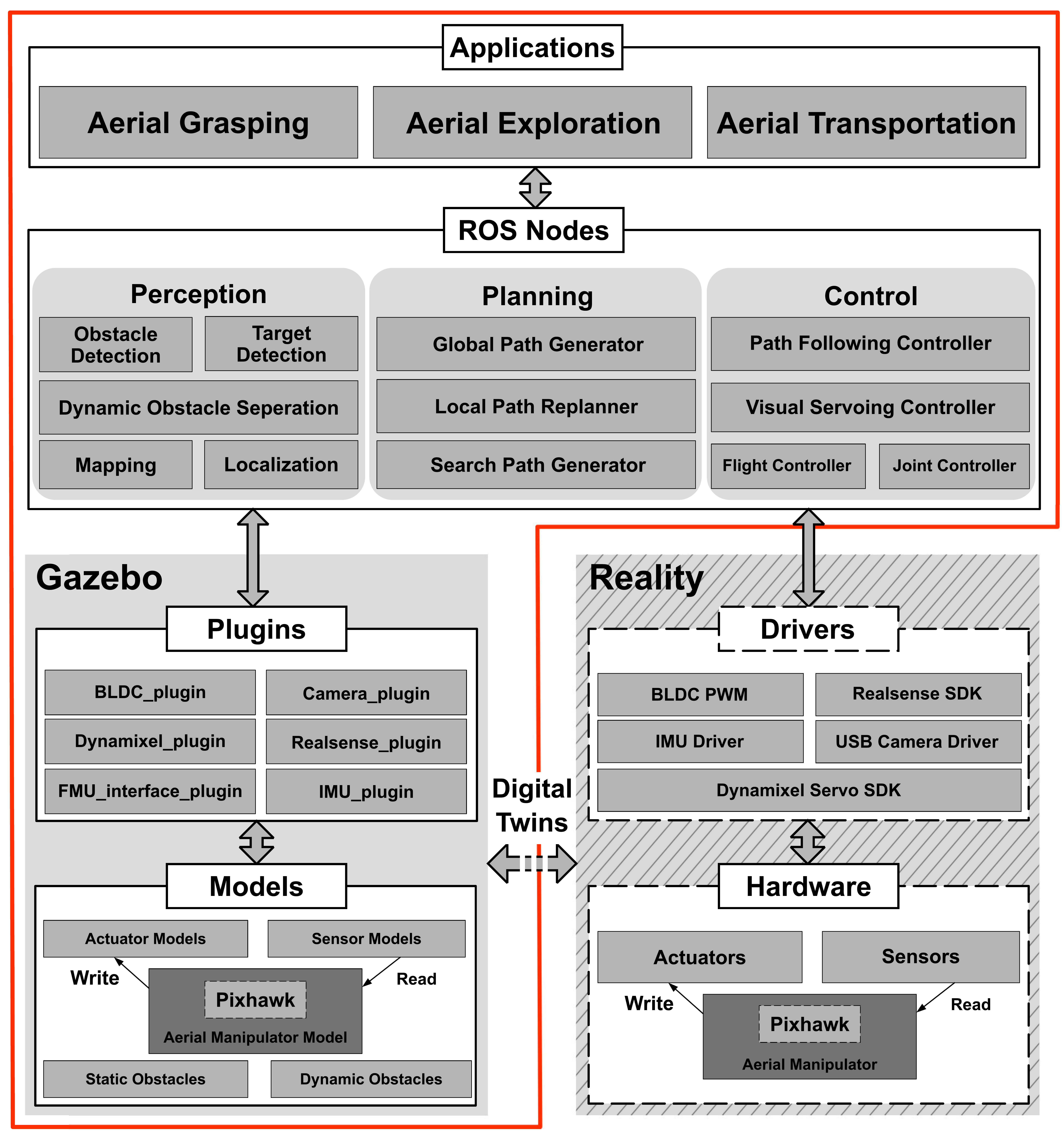}
    \caption{Hierarchical composition structure diagram of the proposed simulation platform.}
    \label{fig:Simulation_Block}
\end{figure}

Fig. \ref{fig:Simulation_Block} shows the diagram of the hierarchical composition structure of the simulation platform. The models and plug-ins in Gazebo environment correspond to the hardware and drivers in real world respectively, forming a digital twins system. The data protocols of the application modules and ROS nodes are compliance to physical systems. The simulation environment is close to the real world, and thus the verification of algorithms on the platform greatly reduce the destructive risks in physical experiments.

\section{Algorithm Architecture for Aerial Manipulation}

\begin{figure}[t]
    \centering
    \includegraphics[width = \linewidth]{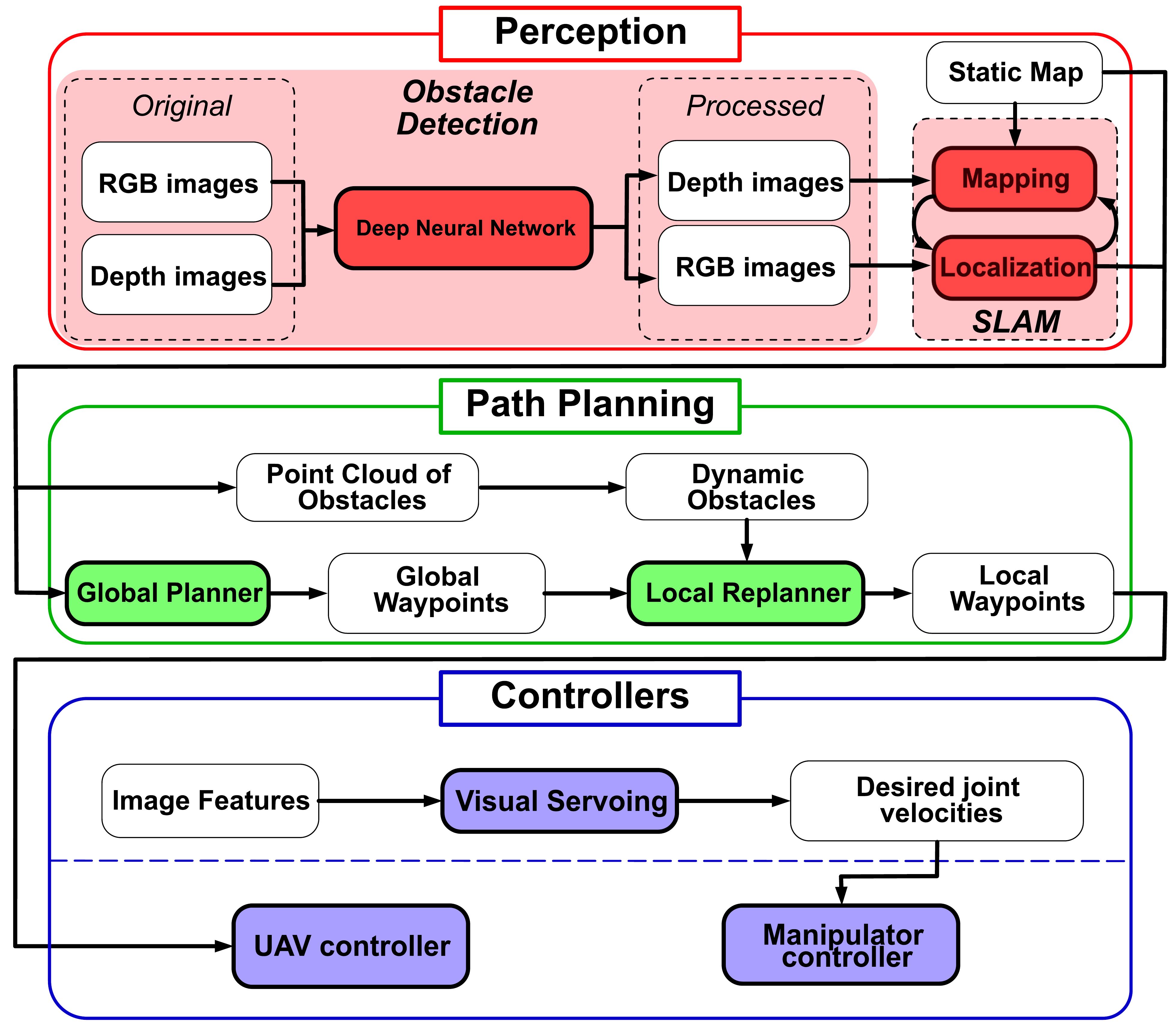}
    \caption{Overview of the proposed algorithm architecture for autonomous aerial manipulation in dynamic environments.}
    \label{fig:architecture}
\end{figure}

Fig. \ref{fig:architecture} illustrates the architecture of the developed algorithm system of AM. 
The perception module builds the environmental map, estimates the AM pose, and extracts dynamic obstacles from static background. Then the perceived information is transferred to the planning module for global and local path generation. Finally, the control module, separated as high and low levels, tracks the planned paths during the whole procedure.
It is noting that the three modules are designed and integrated to make the proposed architecture an efficient and robust system. 
To our knowledge, it is the first time in the literature to provide a complete architecture for autonomous grasping with an AM. 
A more detailed description of each module is given as follows.

\subsection{Perception module}

The perception module needs to not only localize the AM by using on-board sensors, but also detect static obstacles and newly appeared dynamic obstacles. Therefore, this module will consists of two parts, including dynamic simultaneous localization and mapping (SLAM) and dynamic obstacle detection.

\textbf{Dynamic SLAM}: In practical indoor applications, the AM generally can not rely on external localization systems and thus needs on-board sensors to localize itself and perceive surrounding obstacles. According to our previous work \cite{sun2019autonomous}, the visual SLAM (vSLAM) is considered in this paper because of the lightweight and information-rich advantage of cameras.
However, in dynamic environments, the traditional visual SLAM systems like ORB-SLAM2 \cite{mur2017orb} will be seriously degraded in localization accuracy due to the affection of dynamic obstacles. SLAM can even fail when the dynamic obstacle is too close to the camera. 
To eliminate the effect of dynamic obstacles, the deep learning methodology (YOLOv3 \cite{redmon2018yolov3}) is applied; images are pre-processed by extracting a region for potential dynamic obstacles and then removing feature points attached on the moving objects directly. By eliminating the associated 3D moving points, a clean and static feature point cloud map belonging to background only is obtained. The accuracy and robustness of localization in dynamic scenes are improved.

\begin{figure}[t]
    \centering
    \includegraphics[width = \linewidth]{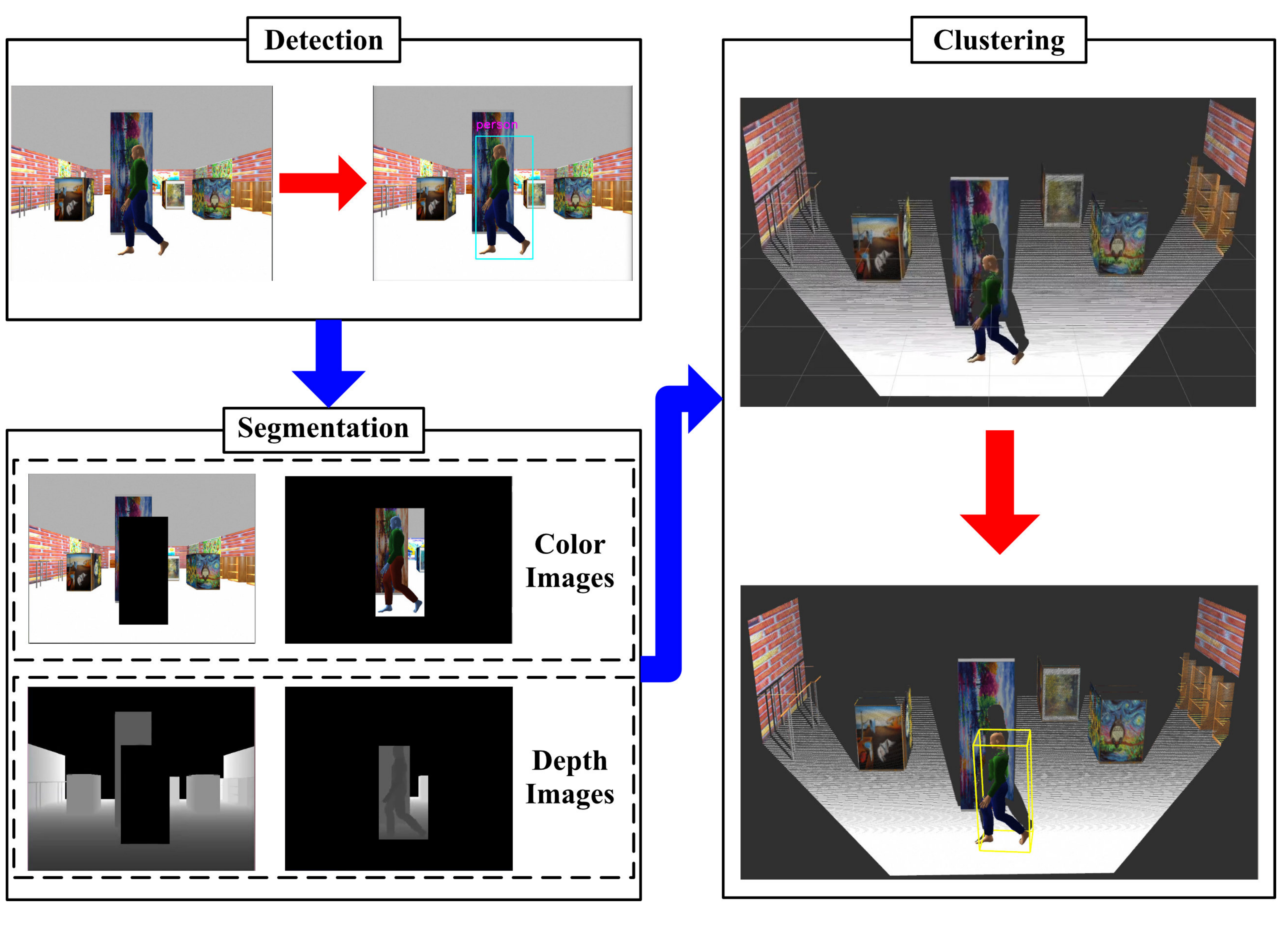}
    \caption{Algorithm pipeline for dynamic pedestrian extraction from each point cloud frame.}
    \label{fig:pipeline_extraction}
\end{figure}

\textbf{Obstacle State Estimation}: New obstacles may appear without prior information from the map in dynamic environments, and thus online obstacle state estimation is inevitable for further autonomous navigation.
By taking advantage of RGB-D camera, the dense voxel map can be obtained with less computational effort compared with stereo camera. 
To estimate the states of dynamic obstacles, the associated point clouds of obstacles are firstly extracted. And then, the ground plane voxels are removed by only considering the height for the convenience of obstacle detection. Without loss of generality, it is assumed that one obstacle occupies most area of one proposed image region which is obtained from deep neural network output. Another assumption is that the detected dynamic obstacle is not occluded, thus the closest point clouds, which is extracted based on image segmentation, are the target point clouds to find. In case of preventing from identifying two nearest point clouds as one object, the point clouds are clustered by utilizing the density-based spatial clustering of applications with noise (DBSCAN) \cite{khan2014dbscan} method. The center points of obstacle point clouds are calculated, and their minimum bounding box are easily extracted by means of the Principal Component Analysis (PCA). The whole algorithm pipeline for obstacle state estimation is illustrated in Fig. \ref{fig:pipeline_extraction}.

\subsection{Planning module}

To guide the AM to approach the target while avoiding dynamic obstacles, an efficient planning module is designed by integrating a global planner, a local planner and a target searching algorithm.

\textbf{Global Planner}: This module aims to obtain a global path without falling into local minimum especially in pre-known cluttered environments. The widely-used sampling-based rapidly exploring random tree (RRT) planner \cite{noreen2016optimal} is applied for its rapid performance to find a feasible path in limited time. As a probabilistic path planner, RRT is a random sampling algorithm that uses incremental growth to solve algebraic constraints and differential constraints. The initial AM location is treated as the tree root node, and a random expanded tree is generated by randomly sampling and adding leaf nodes. When the leaf node in the random tree reaches the target point or enters the target area. A feasible path is obtained by backtracking the random tree to identify edges from the initial point to the target point.
Hereafter, an optimization step is conducted to generate a kinodynamic feasible and obstacle-clear path for reliable movement, which is formulated as a nonlinear quadratic programming problem solved by a quasi-Newton method. 

\textbf{Local Planner}: Due to occlusion and limited viewing angle, the AM may only obtain a crude region but not the exact position of the target. On the other hand, new dynamic obstacles may also appear along the pre-planned path which can lead to collision if the AM merely follows the global path. To address these problems, a region-based local planner developed in previous work \cite{chen2012moving} is applied. The crude region of the target object is defined as a mathematically formulated equation. In addition, the potential field idea is also integrated into the planner. The AM is able to replan a local path even encounter with dynamic obstacles or environmental changes. Compared with traditional methods, the goal of the region-based method is to drive the platform to a certain area instead of a certain precise point while simultaneously avoiding collisions.

\textbf{Target Search}: If the target is not found when the AM moves into the desired region, the camera mounted on the end-effector is then utilized to search the accurate pose of the target. To guarantee finding the target which is visible in limited viewing angles, the camera mounted on the AM needs to enlarge the viewing range as much as possible. Since the target can not be detected in all directions, a search strategy for viewing angle coverage is also developed. 
Considering the workspace of the manipulator, a circular path centering with the coarsely estimated target is set as the desired path for AM to follow, and the end-effector equipped with a camera is driven towards the circular center.
Once the location of the target to be grasped is obtained, it is sent to the control module for further procedure.

\subsection{Control module}

The control module is a fundamental module that make the controlled system converge to desired behaviors like path tracking and grasping. A two-level hierarchical control structure is introduced. The high-level controller is responsible for task oriented control by generating desired states of the AM according to the desired tasks, while the low-level controller ensures good tracking performance in general state space.

\textbf{High-Level Controller}: Since the MAV is a floating and under-actuated platform, the motion of manipulator will cause inevitable disturbance to its aerial base and vice verse. Many existing works try to tackle it by designing separate \cite{kobilarov2014nonlinear} and adjoint controller \cite{lippiello2012cartesian}, aiming to reduce the tracking error of both the MAV and the end effector. However, it is still difficult to eliminate errors thoroughly without high performance sensors and actuators. Our objective is to utilize a simple camera to grasp an object with high success rate. Therefore, a visual servoing controller based on our previous work \cite{quan2018singularity} is applied, and it is able to minimize the coupling effect between the end effector and MAV body from the task level. The controller converts the image feature errors into the rotation signal of the manipulator while the MAV holds the position in front of the target. Once the errors of visual servoing controller converge to zero, the gripper on the manipulator will directly grasp the object.

\textbf{Low-Level Controllers}: The flight control of MAV and the movement control of the manipulator are driven by separate low-level controllers. The MAV uses the traditional cascade PID controller to control its position and attitude. The proposed aerial platform is a coaxial eight-propeller MAV, and thus the actual control objective is the thrust of z-axis in body frame and three-dimensional moments. With the help of inverse kinematics \cite{quan2018singularity}, the relationship between the pose of the end effector and each joint angle can be established. The manipulator also uses PID to control its joint position and speed.

\section{Evaluation Results}

\subsection{Simulation setup}
A series of simulations were performed to illustrate the performance of the proposed platform and algorithm framework. In addition to static obstacles, two simulated pedestrians moving along a predefined trajectory were added in the scene, as shown in Fig. \ref{fig:simulation_environmnet}. The AM was designed to take off from the start location (0,0,0) and grab a cylinder target located at (14.88 m, 0.50 m, 0.83 m) in world frame. To simplify the cylinder detection, an Apriltag was attached on the target; the object detection algorithm without attaching artificial patterns will be integrated into the platform in the near future. Before grabbing, the AM did not know the exact position of the target, and therefore a candidate target region located at (15.0, 0.0, 0.0) was given. Once the AM reaches the candidate region, a circular searching path with the center point around the crude target location is provided. When the target is found, the AM moves to the target front and conduct the final grasping procedure by using the visual servoing controller. The computer running the simulations is equipped with a CPU AMD3900x with 12 cores and 24 threads as well as a graphics card 2080Ti with 11 GB video memory.

\subsection{Evaluation of the perception module}

\begin{figure}[bh!]
        \centering
        \begin{subfigure}[b]{0.5\linewidth}
                \centering
                \includegraphics[width=\textwidth]{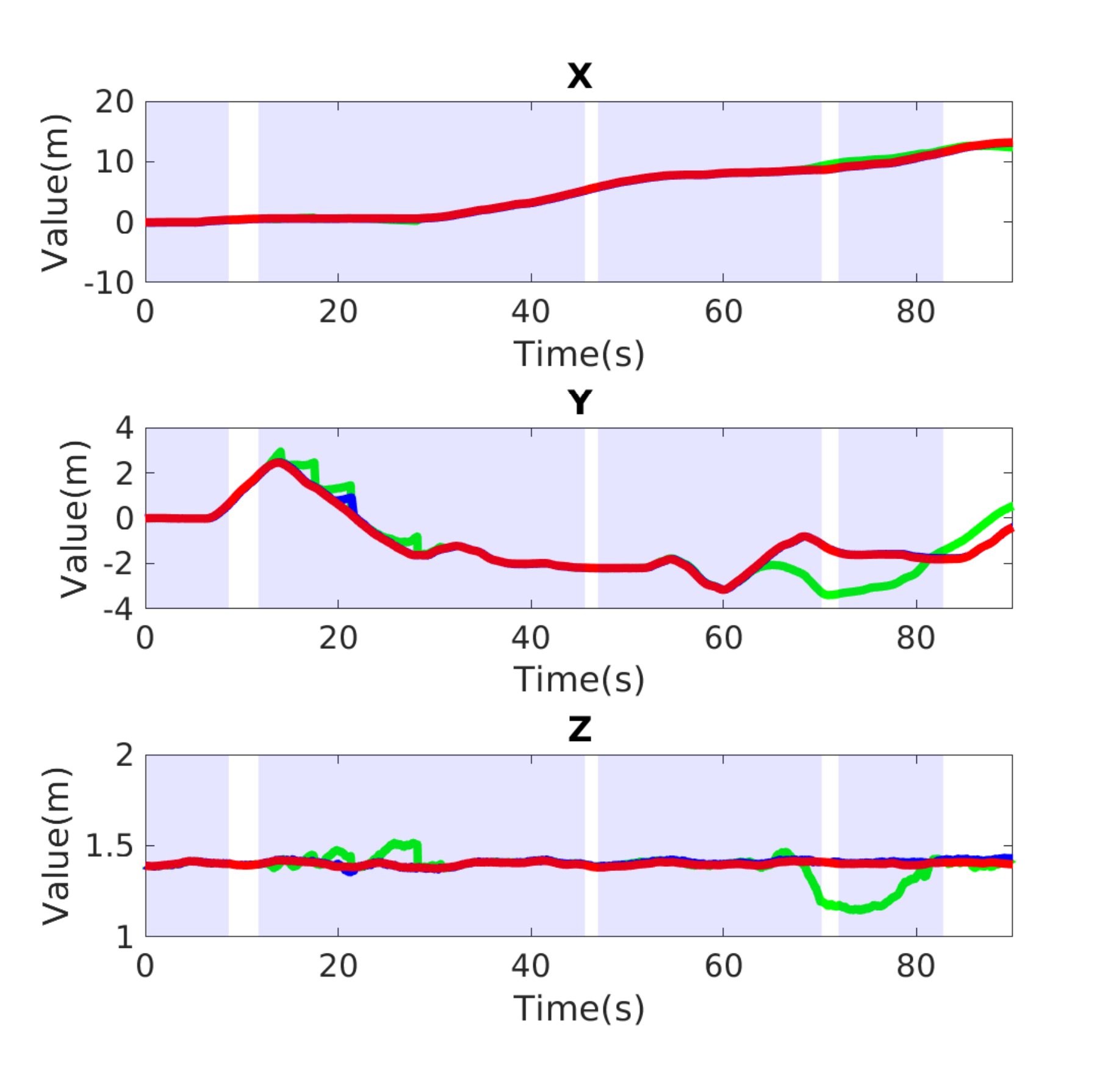}
                \caption{}
                \label{fig:localization_comparison_position}
        \end{subfigure}
        \hfill 
        \begin{subfigure}[b]{0.5\linewidth}
                \centering
                \includegraphics[width=\textwidth]{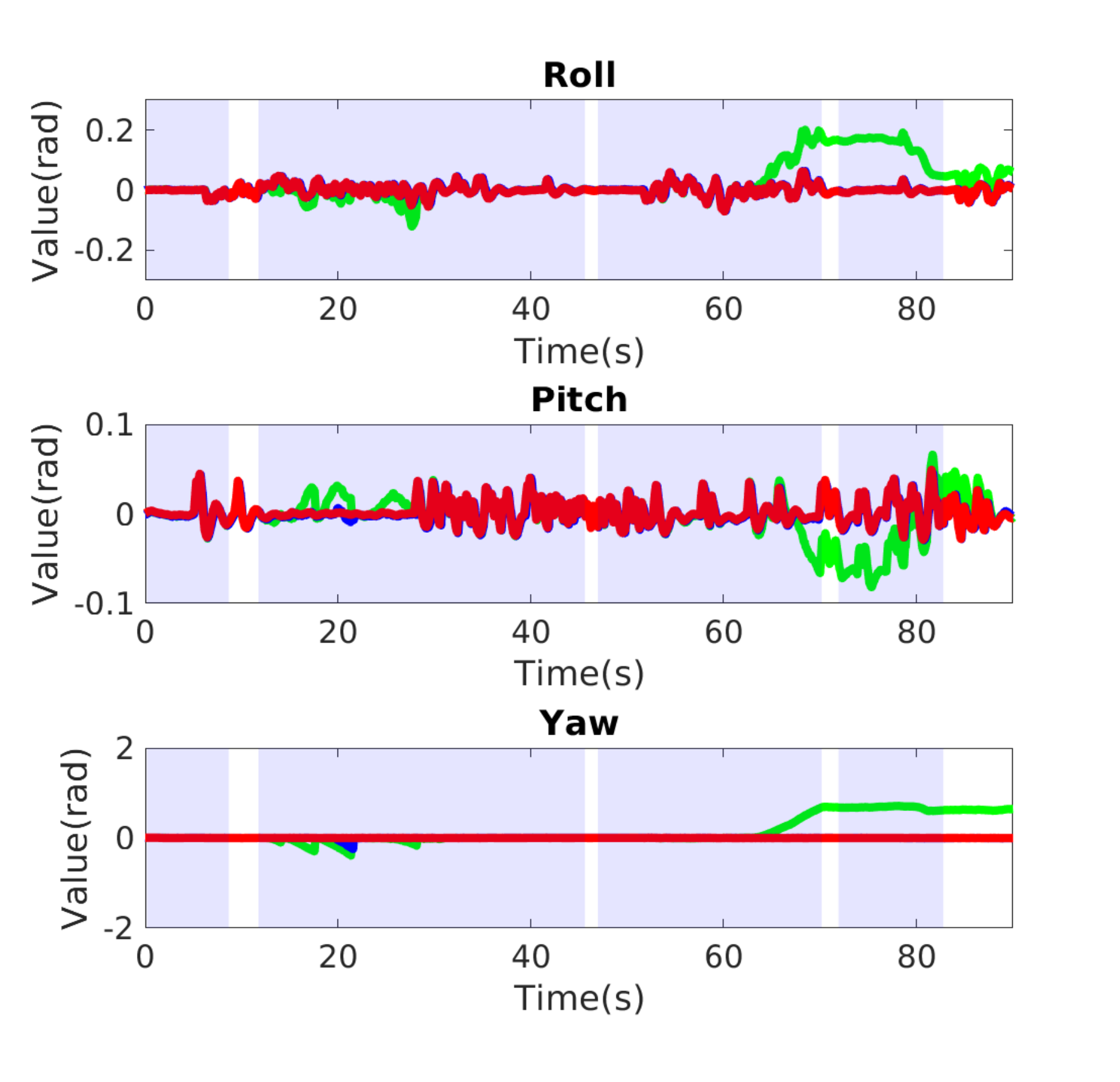}
                \caption{}
                \label{fig:localization_comparison_orientation}
        \end{subfigure}
        \caption{Localization accuracy comparison in (a)position and (b)orientation with/without applying our algorithm.} \label{fig:localization_comparison}
\end{figure}

\begin{center}
\begin{table}[t!]
    \centering
\fontsize{8}{9}\selectfont
    \caption{\label{tab:localization} Comparison of Localization Accuracy}
    \begin{tabular}{p{1.5cm}p{1cm}p{1cm}p{1cm}p{1cm}}
    \toprule
    \toprule
    \multirow{2}[3]{*}{Data} &
    \multicolumn{2}{c}{Average Error} &
    \multicolumn{2}{c}{Max Error}\cr
    \cmidrule{2-3} \cmidrule{4-5}
    & Ours & ORB-SLAM2 & Ours & ORB-SLAM2 \cr 
  
    \midrule
    X/m & \bf{0.007} & 0.046 & \bf{0.086} & 0.778 \cr
    Y/m & \bf{0.004} & 0.065 & \bf{0.022} & 1.585 \cr
    Z/m & \bf{0.005} & 0.015 & \bf{0.017} & 0.192 \cr
    Roll/rad & \bf{0.006} & 0.007 & \bf{0.009} & 0.038 \cr
    Pitch/rad & 0.004 & 0.004 & \bf{0.007} & 0.010 \cr
    Yaw/rad & \bf{0.012} & 0.017 & \bf{0.014} & 0.017 \cr
    \bottomrule
    \bottomrule
   \end{tabular}
\end{table}
\end{center}

Since the AM can only use the onboard RGBD camera to perceive the environment and localize itself, a few patterns are posted in the environment to provide landmarks. The vSLAM algorithm we applied is ORB-SLAM2 for its robust localization performance, and localization accuracy is compared whether YOLOv3 is used to identify and remove dynamic obstacles. As shown in Fig.\ref{fig:localization_comparison}, the red line indicates the ground truth of the RGBD camera pose information, and the blue and green line represent the pose estimation curve with and without YOLOv3 to remove dynamic obstacles, respectively. It is seen that there existed large fluctuations in the pose estimation if dynamic obstacles were not removed; this is because the pose estimation drifts due to the moving obstacles. But after removing the affection of dynamic obstacles, the localization accuracy performed much better. Table. \ref{tab:localization} shows the comparison of localization error data. From the table, the proposed algorithm greatly improves maximum errors which are important for path following and grasping.

\begin{figure}
    \centering
    \begin{subfigure}[b]{0.5\linewidth}
        \includegraphics[width=\textwidth]{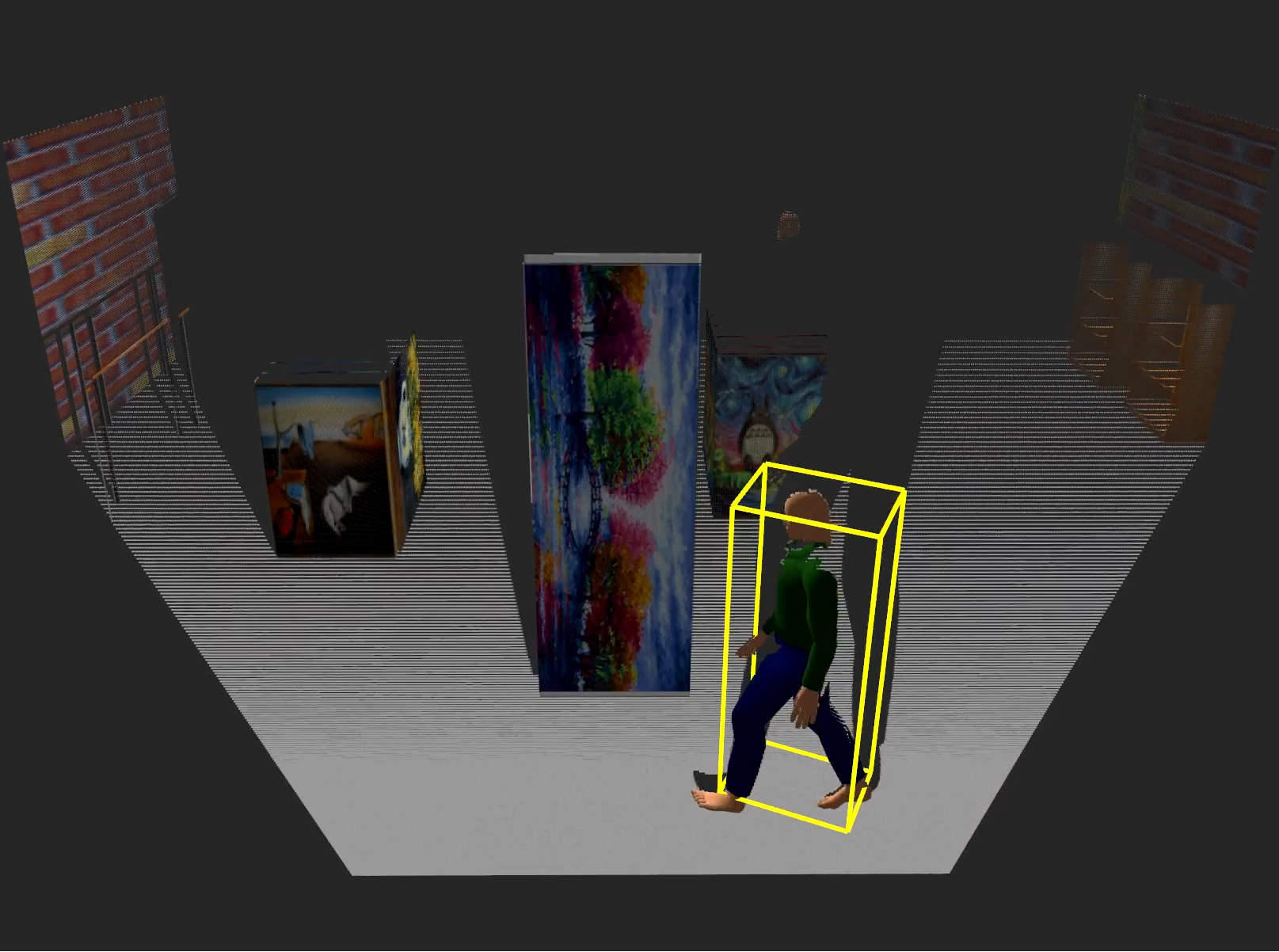}
        \vspace{2mm}
        \caption{}
        \label{fig:obstalce_segmentation}
    \end{subfigure}%
    \hfill
    \begin{subfigure}[b]{0.5\linewidth}
        \includegraphics[width=\textwidth]{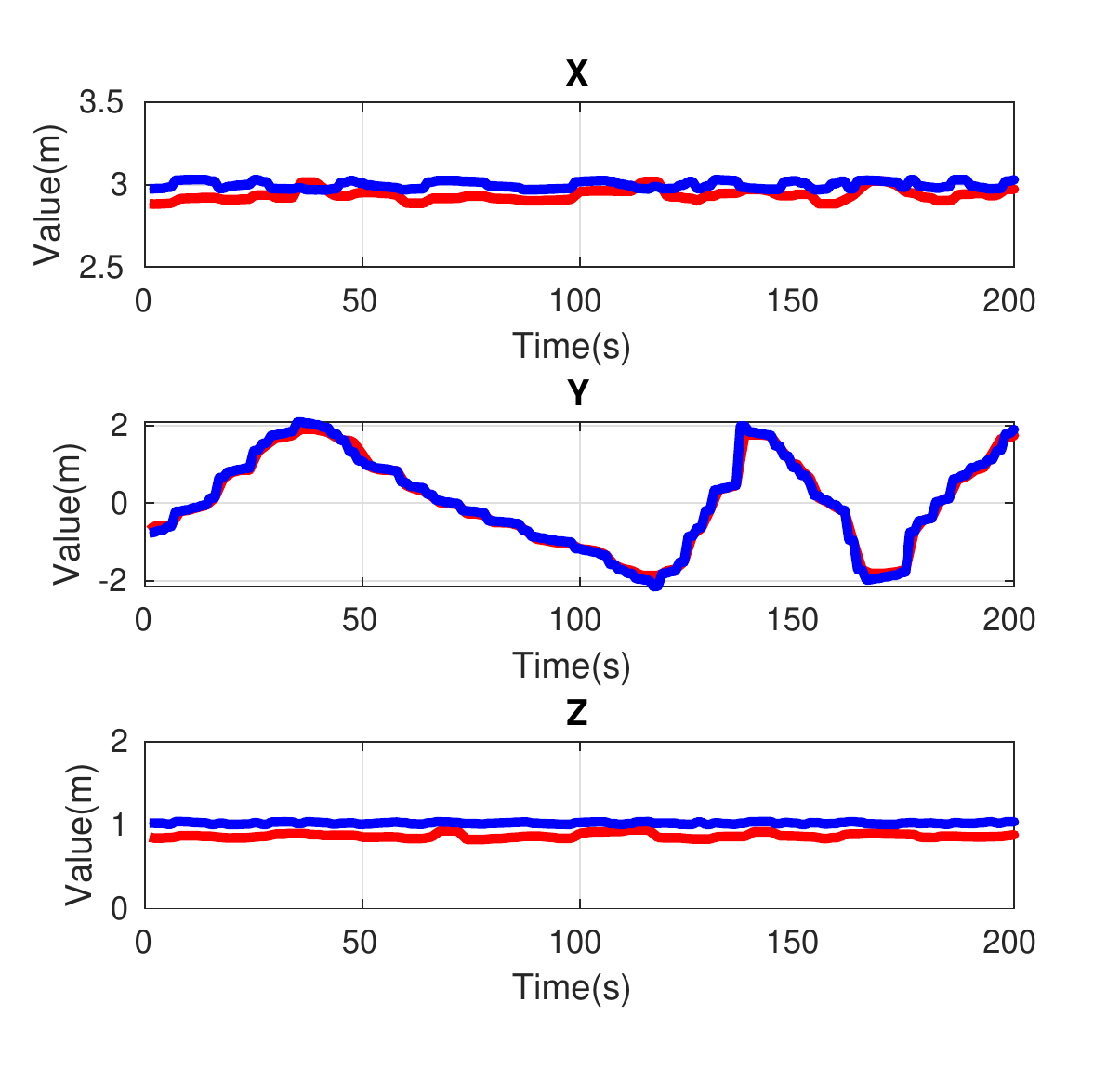}
        \caption{}
        \label{fig:obstalce_pose_estimation}
    \end{subfigure}
    \caption{Pedestrian detection (a) and position estimation data (b). }
    \label{fig:obstalce_estimation}
\end{figure}

In the approaching stage, accurately estimating the pose of dynamic obstacles is also necessary for obstacle avoidance. As shown in Fig. \ref{fig:obstalce_segmentation}, the moving pedestrian extracted from the point cloud is bounded in yellow box. Fig. \ref{fig:obstalce_pose_estimation} shows the pedestrian pose estimation results, where the blue curve indicates the estimated position of the bounding box center and the red curve denotes the ground truth.


\subsection{Evaluation of Planning Module}

\begin{figure}[t]
    \centering
    \begin{subfigure}[b]{0.49\linewidth}
            \centering
            \includegraphics[width=\textwidth, height = 0.65\linewidth]{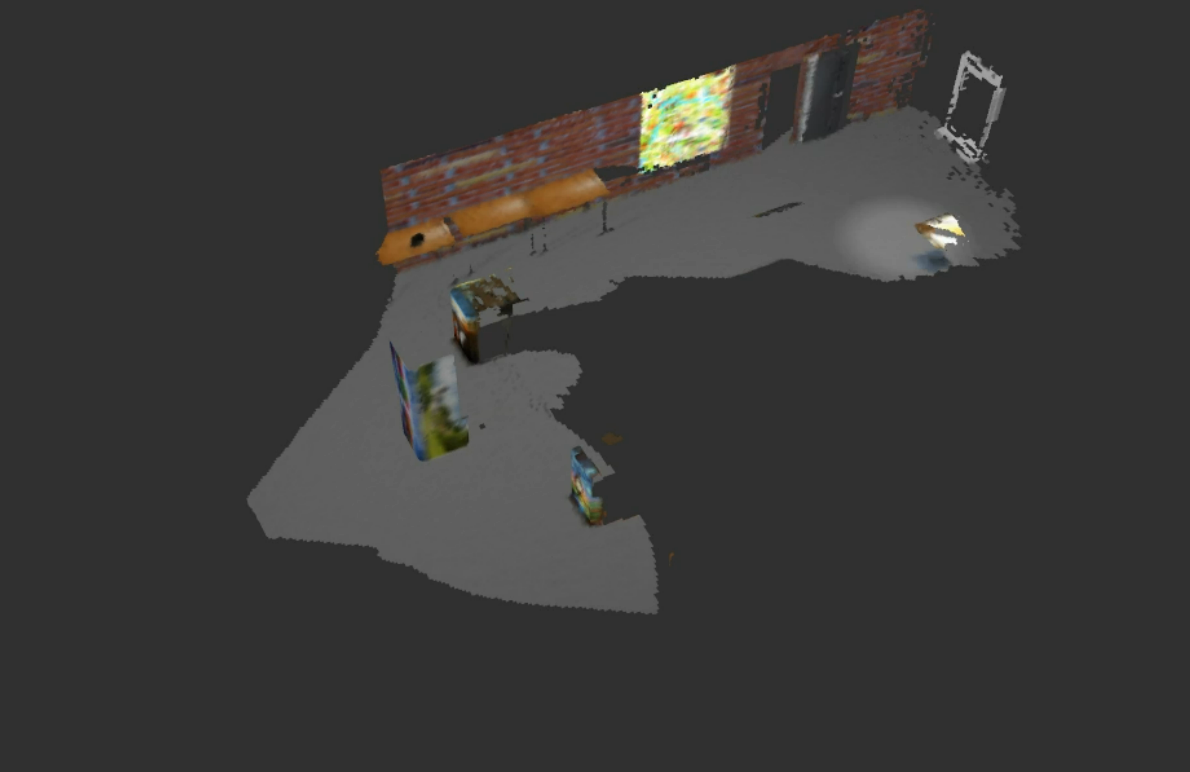}
            \caption{}
            \label{fig:esdf_map_building_1}
    \end{subfigure}
    \hfill 
    \begin{subfigure}[b]{0.49\linewidth}
            \centering
            \includegraphics[width=\textwidth, height = 0.65\linewidth]{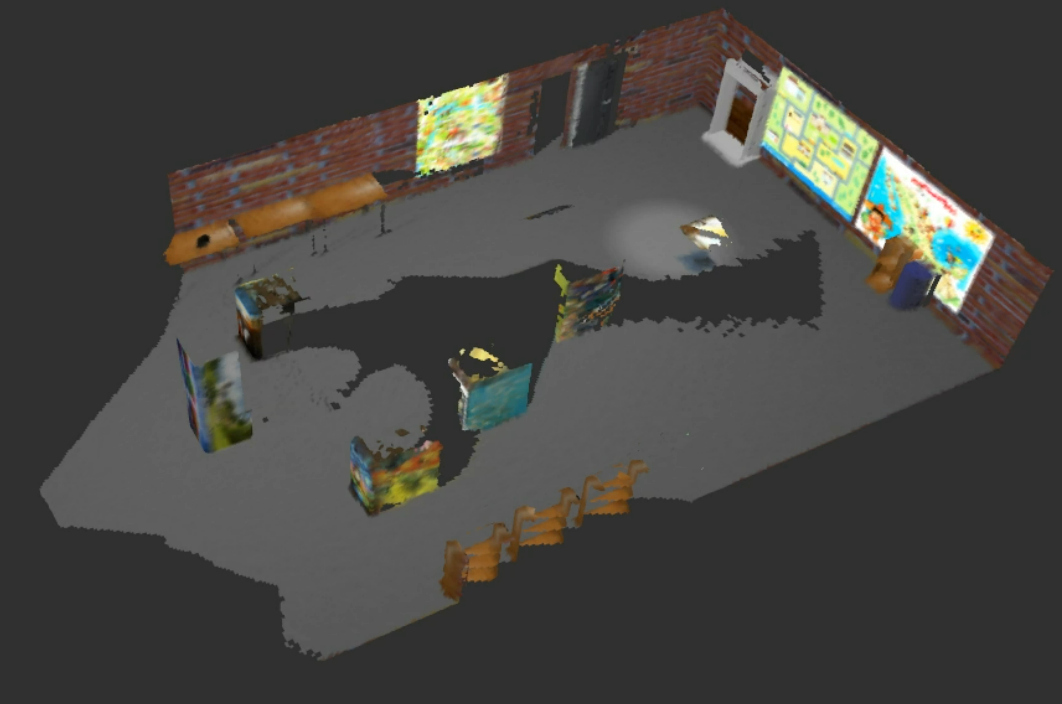}
            \caption{}
            \label{fig:esdf_map_building_2}
    \end{subfigure}
    \hfill 
    \begin{subfigure}[b]{0.49\linewidth}
            \centering
            \includegraphics[width=\textwidth, height = 0.65\linewidth]{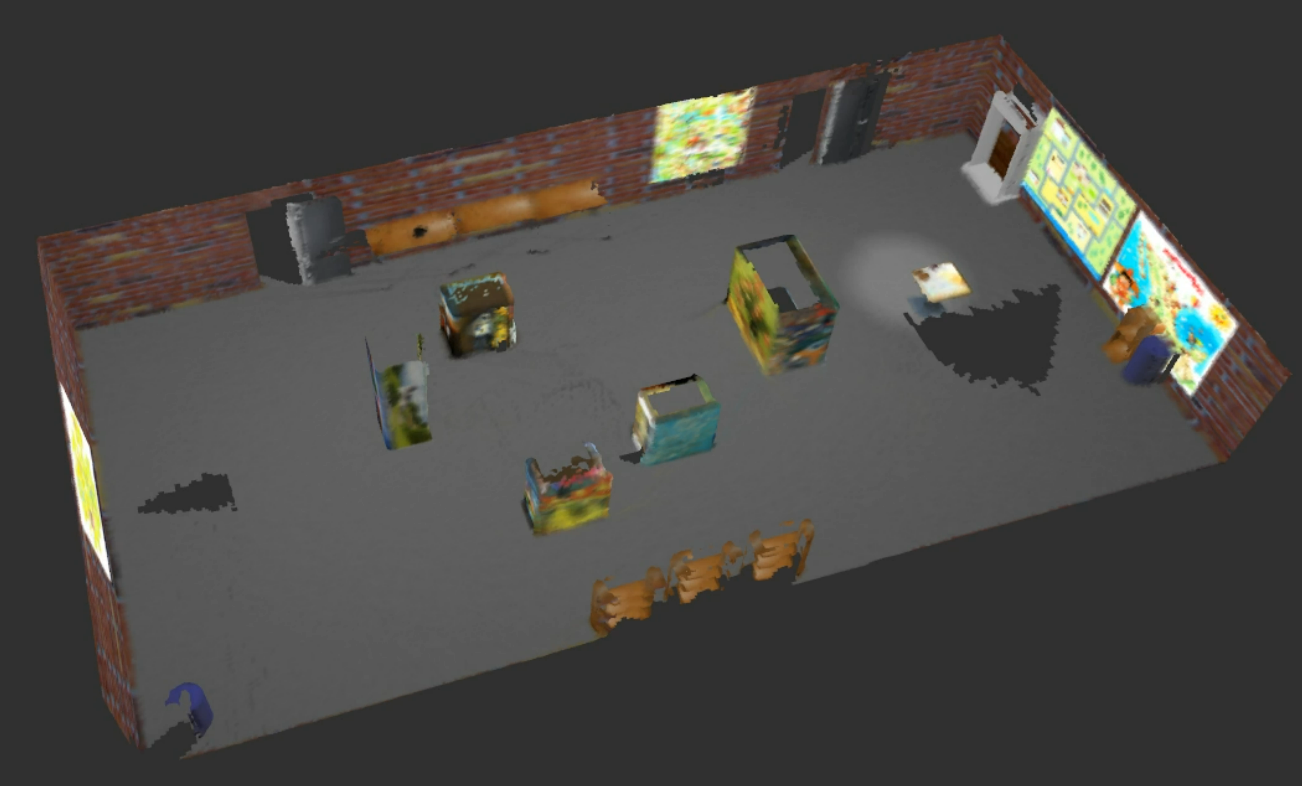}
            \caption{}
            \label{fig:esdf_map_building_3}
    \end{subfigure}
    \hfill 
    \begin{subfigure}[b]{0.49\linewidth}
            \centering
            \includegraphics[width = \linewidth, height = 0.65\linewidth]{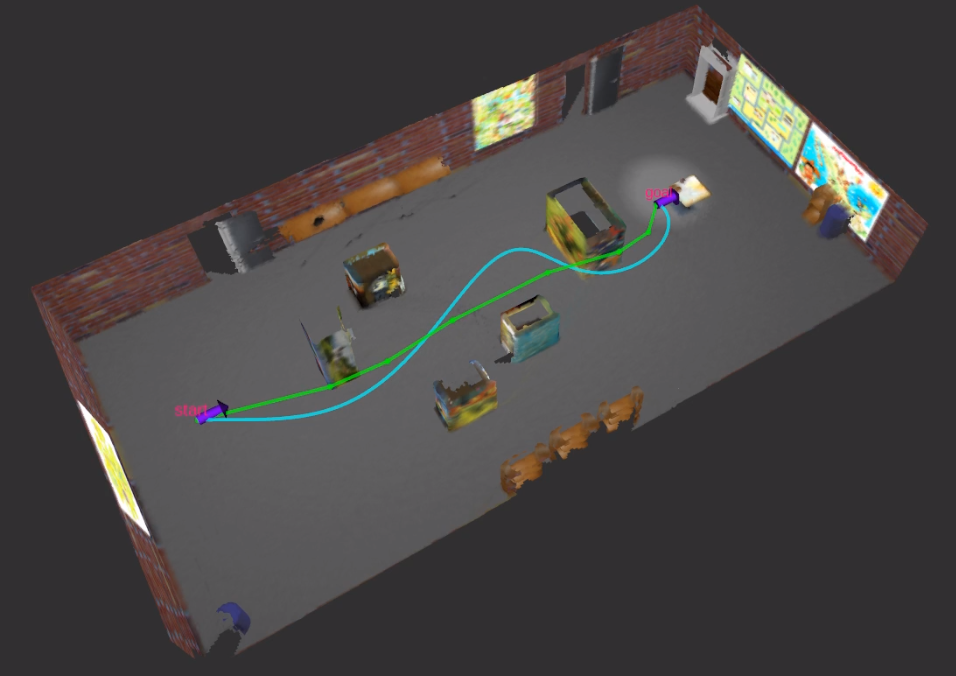}
            \caption{}
            \label{fig:global_path_generation}
    \end{subfigure}
    \caption{(a)-(c)ESDF map building process; (d)Globally planned path in pre-build ESDF map}
    \label{fig:global_path}
\end{figure}

To evaluate the planning module, an ESDF map of the environment without dynamic obstacles was first built with the perception module, as illustrated in Fig.\ref{fig:esdf_map_building_1}-Fig.\ref{fig:esdf_map_building_3}. By setting the start location as well as the crude location of the target, a feasible path as shown with the green curve in Fig. \ref{fig:global_path_generation}, was first generated, based on which a kinodynamic feasible and smoothed path was generated by utilizing the proposed planning module. The final path is shown with the teal curve in Fig. \ref{fig:global_path_generation}. The average computational time is 324ms for RRT* path generation and 10.78s for optimized path generation, when each voxel size is set to 0.05m and the whole distance between the start point and the end point is 13.86m. 

\subsection{Evaluation of Control Module}

\begin{figure}[h]
    \centering
    \includegraphics[width = \linewidth]{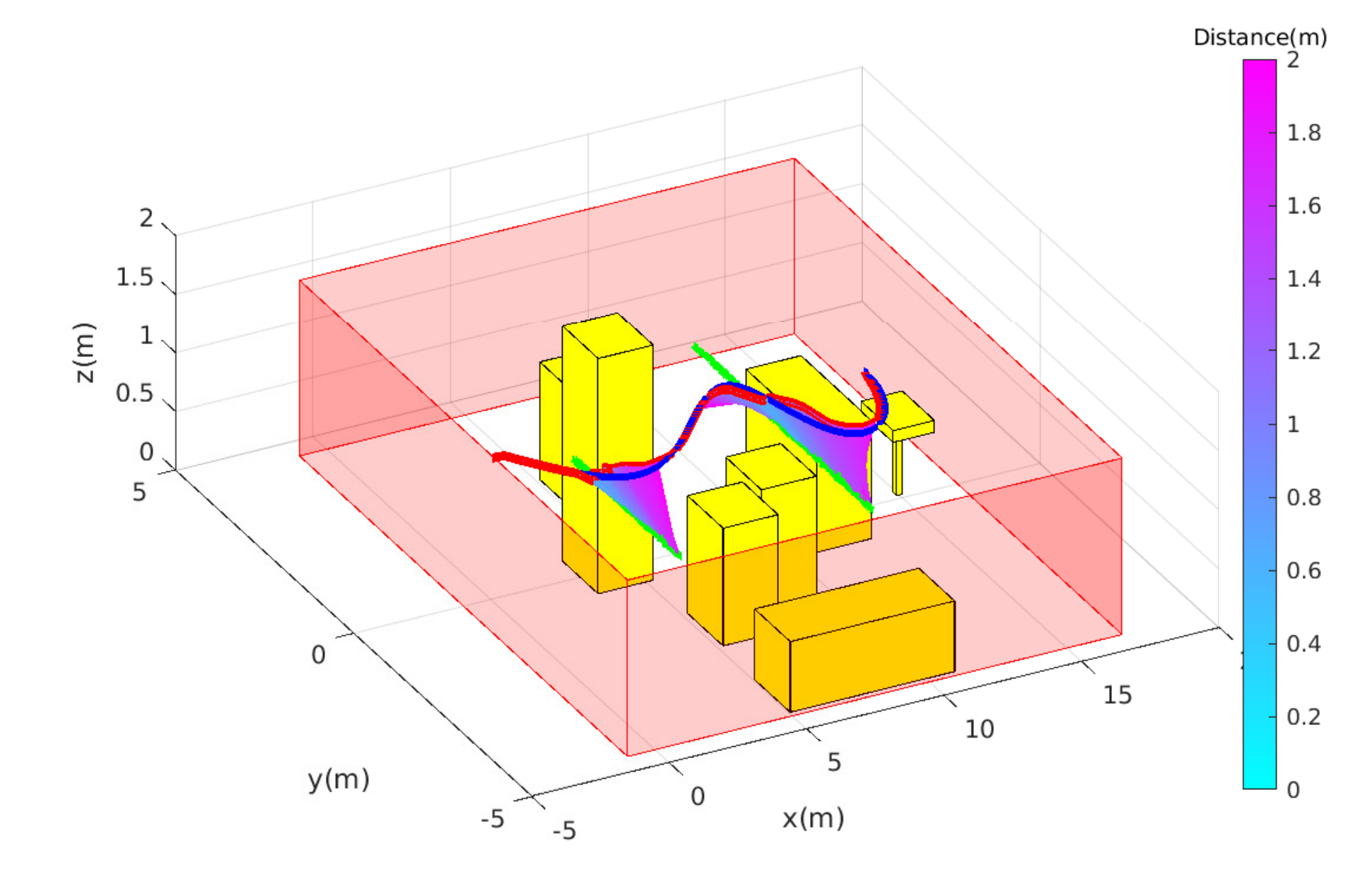}
    \caption{Path following and obstacle avoidance result with two pedestrians in the environment.}
    \label{fig:path_following}
\end{figure}

Fig.\ref{fig:path_following} illustrates the path following process.
The blue curve denotes the global path while the red curve represents the actual path of aerial platform. 
Yellow cubes denote the static obstacles, and the outer transparent red and uncovered cube represents the wall. Two straight lines in the middle of the scene, depicted in green, represent the movement trajectory of the two pedestrians. 
The distance that triggers the generation of a repulsive force by an obstacle is set by 0.8m in our simulation environment.
And the Euclidean distance between the AM base and pedestrians along the actual trajectory is visualized with different colors illustrated in the color-bar aside. 
By applying our local planner, the AM is pushed away by a virtual repulsive force derived from the encountered obstacle. 

\begin{figure}
        \centering
        \begin{subfigure}[b]{0.5\linewidth}
                \centering
                \includegraphics[width=\textwidth]{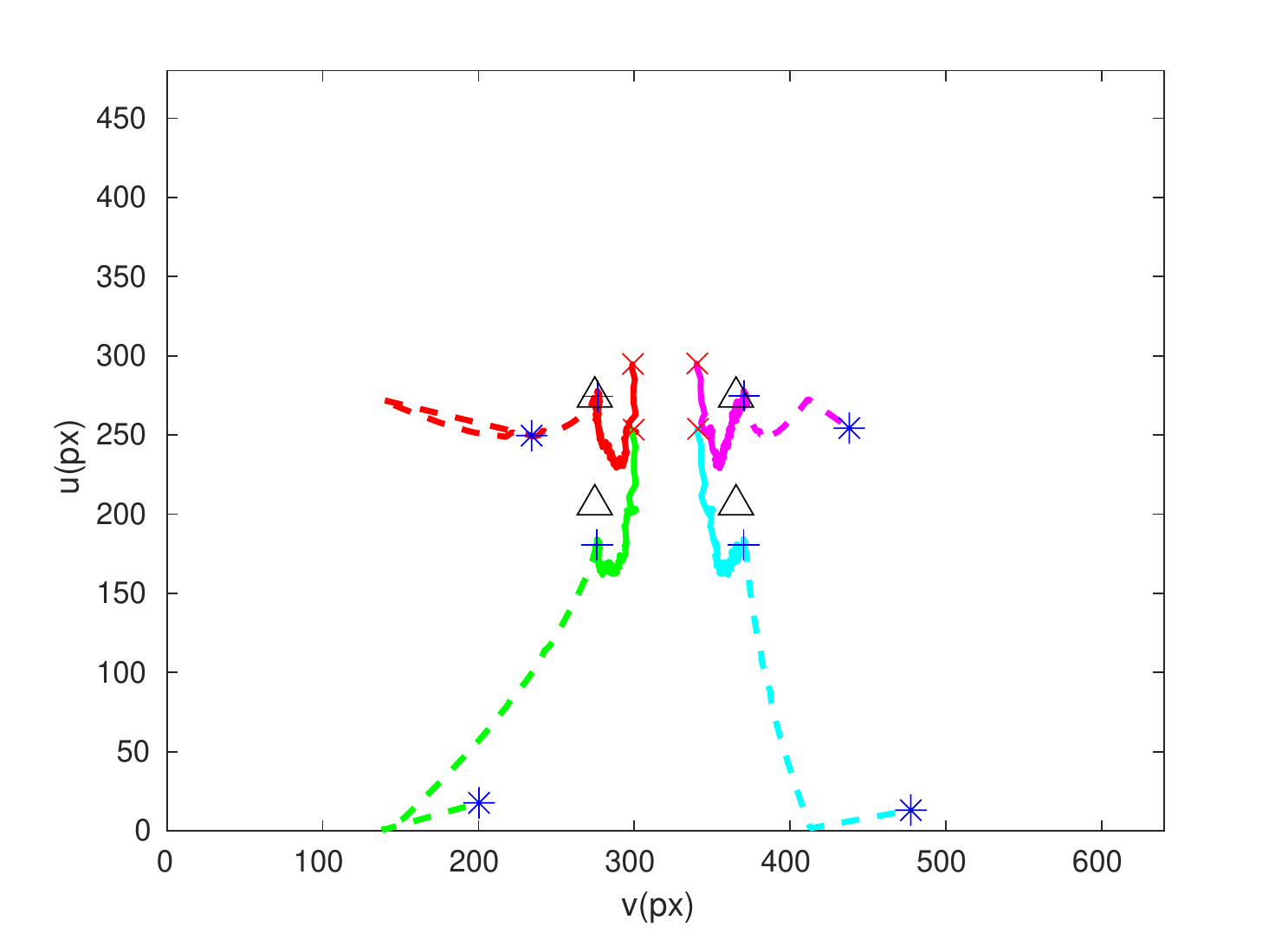}
                \caption{}
                \label{fig:vs_points_trj}
        \end{subfigure}
        \hfill 
        \begin{subfigure}[b]{0.5\linewidth}
                \centering
                \includegraphics[width=\textwidth]{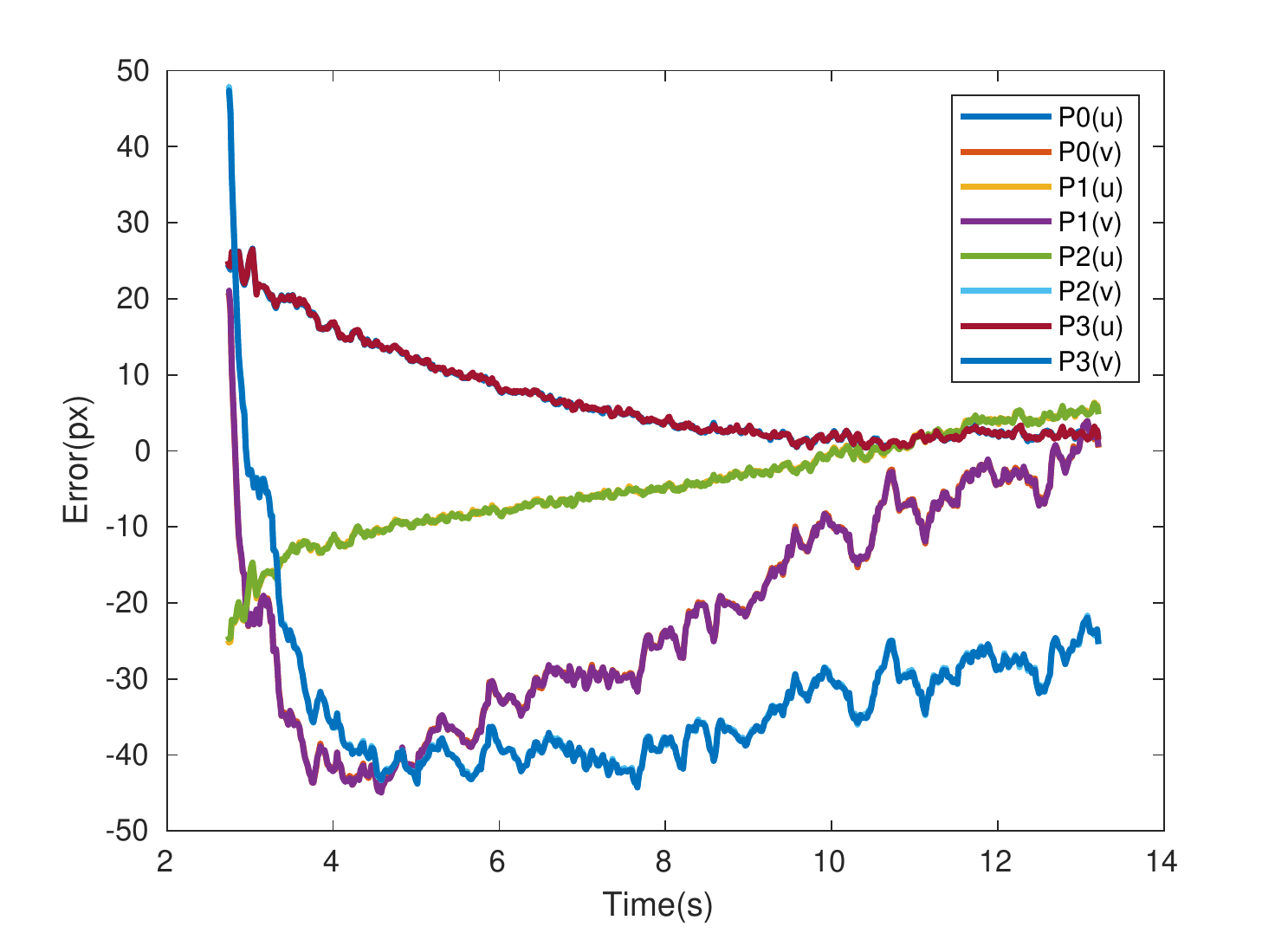}
                \caption{}
                \label{fig:vs_points_err}
        \end{subfigure}%
        \hfill 
        \begin{subfigure}[b]{0.5\linewidth}
                \centering
                \includegraphics[width=\textwidth]{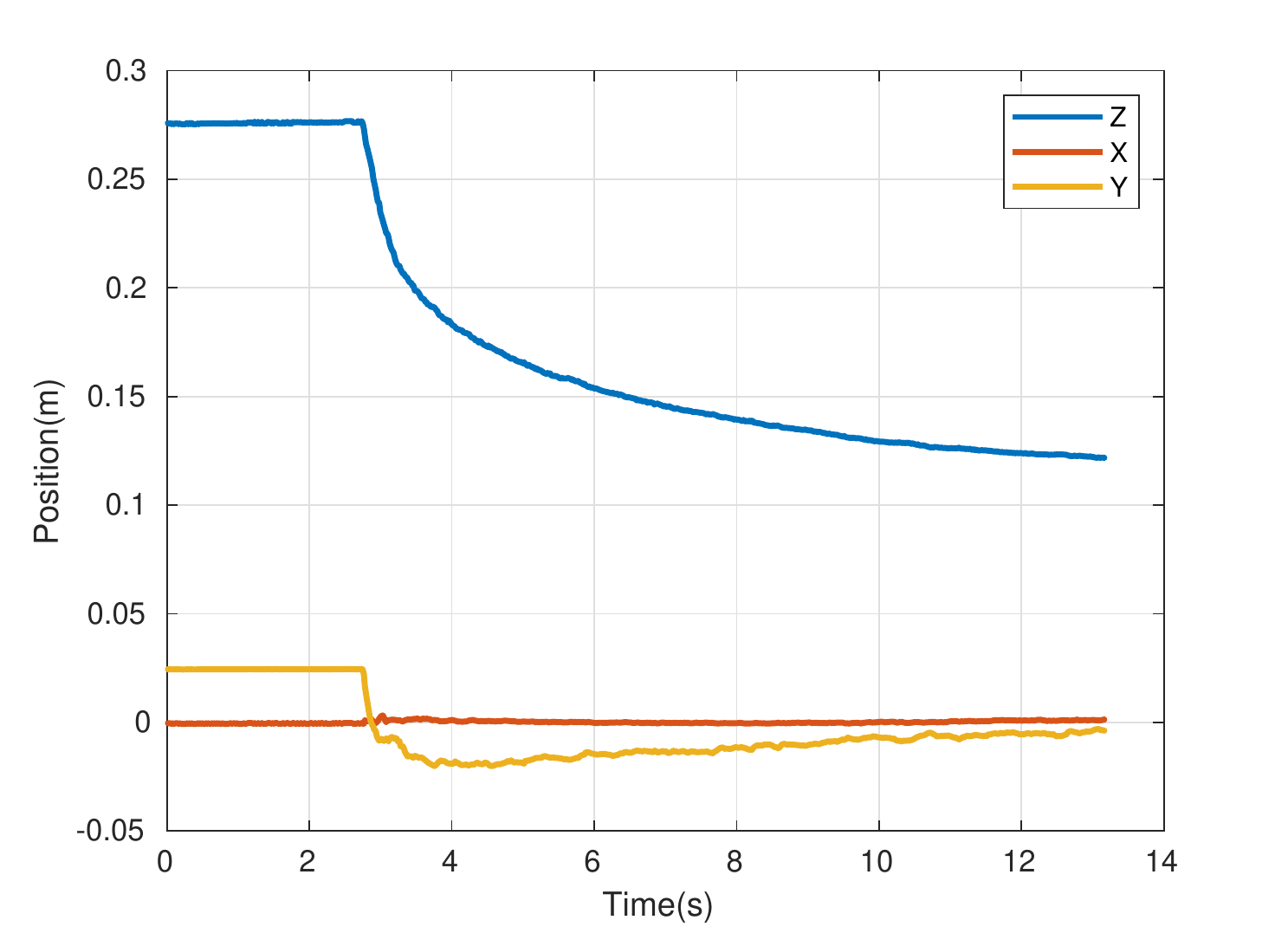}
                \caption{}
                \label{fig:vs_position_err}
        \end{subfigure}%
        \hfill 
        \begin{subfigure}[b]{0.5\linewidth}
                \centering
                \includegraphics[width=\textwidth]{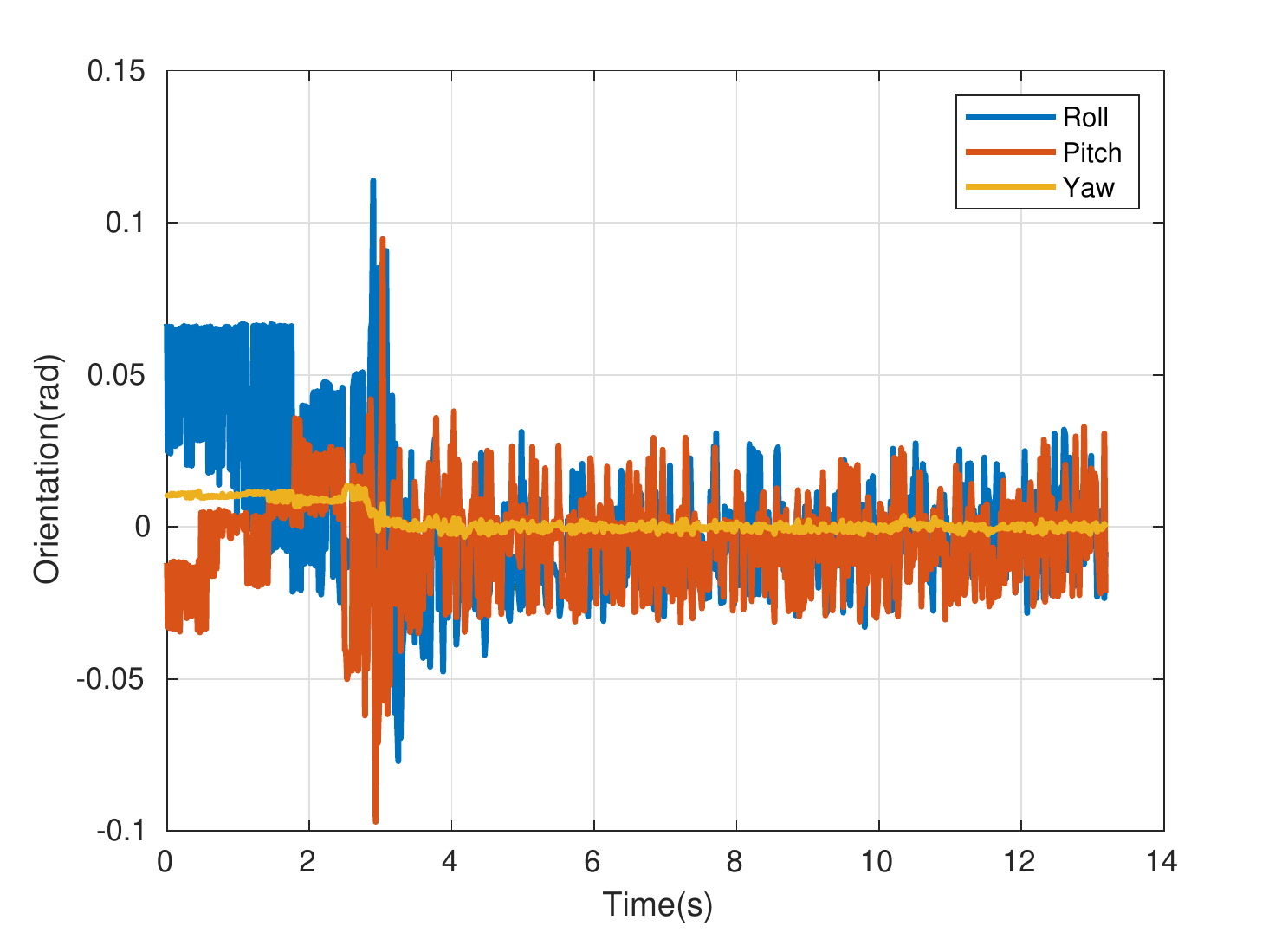}
                \caption{}
                \label{fig:vs_orientation_err}
        \end{subfigure}
        \caption{Results of the visual servoing-based aerial grasping control. (a) Trajectories of the four feature points during the control; (b) Errors in image plane; (c) Position and (d) Orientation errors.} \label{fig:visual_servoing_result}
\end{figure}

To further evaluate the visual servoing control algorithm, aerial grasping simulation was performed.
Because the camera for visual servoing is installed at the end of the manipulator, the output of the controller is easily transformed into the velocity of the end-effector by utilizing a constant transformation from the camera frame to the end-effector.
Fig. \ref{fig:vs_points_trj} illustrates the trajectories of the four corner feature points on image plane in pixel from the beginning of visual servoing to the final grasp of the target object. In Fig. \ref{fig:vs_points_trj}, the triangles indicate the desired positions of feature points; the red cross and blue + indicate the positions of each feature point when the visual servoing starts and ends, respectively; the blue * indicates the position when the target is finally captured. The curve in Fig. \ref{fig:vs_points_err} shows the change of the normalized coordinates of the four feature points during the visual servoing process.
The target pose with respect to the camera is shown in Fig. \ref{fig:vs_position_err} and \ref{fig:vs_orientation_err}. It is seen that the camera converged to the desired position in front of the target by 0.12 m, and the visual servoing stage terminated and triggered the grasping process. Finally, the target object was successfully grasped. As tested on our simulated system, the success rate of grasping is 70\% (21/30). The reason for grasping failure is that the planning of the manipulator is achieved in joint space instead of the task space. The contact detection is not considered, and we will tackle this issue in our future work.

\section{Conclusion}
In this paper, a new modular simulation platform is designed to evaluate aerial manipulation related algorithms. This system helps to prevent destructive results caused by directly deploying the algorithms to real systems. In addition, to realize a fully autonomous aerial grasping, a series of algorithm modules constituting a complete workflow are designed and integrated in the simulation platform, including perception, planning and control modules. This framework empowers the AM to autonomously grasp remote targets without colliding with surrounding obstacles relying only on on-board sensors. Benefiting from its modular design, this software architecture can be easily extended with additional algorithms. Several evaluations are performed to illustrate the effectiveness of the proposed platform and algorithms. Our future work will include real-world experiments by integrating the simulation platform and our experimental aerial manipulator in GPS-denied indoor and outdoor dynamic environments, and the whole simulation system will be open source.

\medskip

\bibliographystyle{IEEEtran}
\bibliography{manuscript_iros}

\end{document}